\pdfoutput=1
\documentclass[11pt,a4paper]{article}
\usepackage{emnlp2021}
\usepackage{times}
\usepackage{latexsym}
\usepackage{paralist}
\usepackage[shortlabels]{enumitem}
\usepackage{graphicx}

\usepackage[footnoteinside=false]{mdframed}
\usepackage[colorinlistoftodos,prependcaption,textsize=tiny]{todonotes}
\usepackage{cleveref}

\usepackage{amssymb}

\newcommand{\checkbox}{$\square$}%

\newlist{checklist}{enumerate}{3}
\setlist[checklist,1]{label={\arabic*.~\checkbox},ref=\arabic*,leftmargin=3em}
\setlist[checklist,2]{label={\thechecklisti.\arabic*.~\checkbox},ref={\thechecklisti.\arabic*}}
\setlist[checklist,3]{label={\thechecklistii.\arabic*.~\checkbox},ref={\thechecklistii.\arabic*}}

\usepackage{microtype}
\usepackage{babel}

\providecommand{\e}[1]{{\protect\color{black}{#1}}}

\title{`Just What do You Think You're Doing, Dave?'\thanks{\hspace*{1mm}
    A line from the Stanley Kubrick film \textit{2001: A Space Odyssey}.}\\ A Checklist for Responsible Data Use in NLP}

\author{Anna Rogers  \\
  Center for Social Data Science \\
  University of Copenhagen \\
  Denmark \\
  \url{arogers@sodas.ku.dk} \\\And
  Timothy Baldwin \\
  CIS\\
  The University of Melbourne\\
  Australia\\
  \url{tb@ldwin.net} \\\And
  Kobi Leins \\
  King's College / London \\
  via Australia\\
  \url{kobi.leins@kcl.ac.uk} \\  
  }

\date{}

\begin{document}
\maketitle
\begin{abstract}
A key part of the NLP ethics movement is responsible use of data, but exactly  what that means or how it can be best achieved remain unclear. This position paper discusses the core legal and ethical principles for collection and sharing of textual data, and the tensions between them. We propose a potential checklist for responsible data (re-)use that could both standardise the peer review of conference submissions, as well as enable a more in-depth view of published research across the community. Our proposal aims to contribute to the development of a consistent standard for data (re-)use, embraced across NLP conferences.
\end{abstract}

\section{Introduction}

In NLP, as in other research areas that are heavily data-driven, it is impossible to overstate the significance of the creation, collection, storage, and use of data \e{\cite{PaulladaRajiEtAl_2020_Data_and_its_discontents_survey_of_dataset_development_and_use_in_machine_learning_research}}. Machine learning models are fundamentally a way of storing (and ideally generalising over) training data in the form of latent representations. The vast majority of NLP tasks involve supervised training of some sort%
. Ideally, the training data would come from the real-world, be of high quality, diverse, available in large quantities, obtained with \e{revocable} consent for specific use cases, and in compliance with licensing and other legal obligations, as well as broader ethical considerations. %

Unfortunately, many of these considerations are often overlooked, to the potential detriment of \e{both} quality and integrity of the \e{resulting datasets and models}. 
A key issue is that the legal frameworks vary from jurisdiction to jurisdiction, as do the workplace cultures and norms around \e{research institutions}.%
We \e{may} not even have established definitions for commonly used terms like ``fairness'' \cite{MulliganKrollEtAl_2019_This_Thing_Called_Fairness_Disciplinary_Confusion_Realizing_Value_in_Technology,XiangRaji_2019_On_Legal_Compatibility_of_Fairness_Definitions} %
and ``bias'' \cite{BlodgettBarocasEtAl_2020_Language_Technology_is_Power_Critical_Survey_of_Bias_in_NLP}. %
The result is %
differing expectations and standards, \e{which makes it harder for fellow researchers to understand the terms of (re-)use of new resources and models, and also complicates peer review.} %

Experts in biomedical data research have concluded that \e{``truly informed consent ... requires} (1) comprehensive disclosure of informational risks to participants, (2) independent governance entities, and (3) data sharing policies that offer guidance for physicians and researchers'' \cite{Mittelstadt_2019_Ethics_of_Biomedical_Big_Data_Analytics}. %
The AI community has already started to work on the latter \cite{Mantelero_2018_AI_and_Big_Data_blueprint_for_human_rights_social_and_ethical_impact_assessment,DuanEdwardsEtAl_2019_Artificial_intelligence_for_decision_making_in_era_of_Big_Data_-_evolution_challenges_and_research_agenda,SielemannHafnerEtAl_2020_reuse_of_public_datasets_in_life_sciences_potential_risks_and_rewards,2017_Asilomar_AI_Principles}. Unfortunately, the codes of conduct and promulgation of ethics by industry, government, and academia remain highly contextual, and have ``substantive divergence in relation to how these principles are interpreted, why they are deemed important, what issue, domain or actors they pertain to, and how they should be implemented'' \cite{JobinIencaEtAl_2019_global_landscape_of_AI_ethics_guidelines}. %
Further, a more cynical, albeit well-founded view, is that much of the flurry of activity around codes of conduct and ethics has been to avoid regulation, and is heavily driven by industry \cite{MetcalfMossEtAl_2019_Owning_Ethics_Corporate_Logics_Silicon_Valley_and_Institutionalization_of_Ethics}.

In NLP, the Association for Computational Linguistics (ACL) has adopted the ACM code of ethics \cite{GotterbarnBrinkmanEtAl_2018_ACM_Code_of_Ethics_and_Professional_Conduct}, conducted a series of workshops \cite{HovySpruitEtAl_2017_Proceedings_of_First_ACL_Workshop_on_Ethics_in_Natural_Language_Processing,AlfanoHovyEtAl_2018_Proceedings_of_Second_ACL_Workshop_on_Ethics_in_Natural_Language_Processing}, and, most recently, implemented ethics reviews and submission guidelines.\footnote{\e{The NAACL guidelines (\url{2021.naacl.org/ethics/faq}) were subsequently adapted by ACL-IJCNLP and EMNLP 2021.}} But we are far from a set of rules that would account for all the diverse types of NLP data and applications. \e{Initial discussions after the first round of ethics review in NAACL 2021 highlighted many disagreements.}

We submit that the community needs to work towards a common standard for what constitutes responsible use of data, what review mechanisms should be in place to ensure it, and what should happen when that standard is not upheld. Ideally, the discussion would advance collectively and with representation across different sub-populations of the field, \e{as well as other stakeholders}. Hopefully this process would result in more comprehensive disclosure, independent governance structures, and %
clear\e{er} policies that reflect a common purpose by the NLP community. %

To this end, we analyse the core legal and ethical principles of data collection, and distill best-practice recommendations into a (non-exhaustive) checklist for potential adoption/adaptation by NLP conferences. Above all, we hope that this paper will advance discussion towards a single standard of responsible data (re-)use in NLP.%

\section{Key Principles of Data Collection}

We do not yet have unified standards for responsible data use, but there are several guiding principles upon which we  widely agree. This section briefly discusses such principles and the interactions between them. 

\subsection{Copyright and Terms of Use}
\label{sec:copyright}

In general, any data that is not in the public domain %
is considered to be the property of the individual or entities that produced it, and explicit permission is needed to collect/distribute it. It is incumbent upon researchers to ensure that they do not violate the law in undertaking their work. But even this law is not straightforward, because different jurisdictions have different copyright requirements. In 2021, e.g, US-based researchers cannot publish resources based on Hemingway's works, as copyright expires 70 years after the creator's death, but Japan-based researchers can: for them it is 50 years. 

In addition to protection by copyright law, data is frequently protected by specific terms of services (``ToS'') of the hosting platform, such as Yelp or Twitter. While copyright applies automatically, ToS have to be presented to users before they engage with the content (whether as a data generator or collector), and have to be explicitly agreed to. That said, they are often long, impenetrable, and many users (even researchers) give consent without reading them \e{\cite{ObarOeldorf-Hirsch_2020_biggest_lie_on_Internet_ignoring_privacy_policies_and_terms_of_service_policies_of_social_networking_services}}. But once ToS are accepted, any violations may have potential legal consequences. Most platforms have strict limitations on who can use what data for what purposes, and any use not in compliance with the ToS is probibited -- \e{with the complication that most such services operate internationally, with possibly different versions of ToS in different locations, and with different applicable international, national and regional laws. ToS may also change between the time of data collection and resource publication, and at this point there is no easy answer to the question of which version should apply.}\footnote{\e{Generally, law does not apply retroactively \cite{KryvoiMatos_2021_Non-Retroactivity_as_General_Principle_of_Law}. Given the bargaining power disparity between the platform and their users, whether the platforms are free to unilaterally change their terms with retroactive effect is a highly complex question concerning many fields of law, including unfair competition and consumer protection. At present, a legally accurate assessment for a particular case would have to start from determining the applicable local law (e.g. the law of Netherlands), and how it treats the particular type of contract (e.g. as service contract).}}

Different countries may also limit specific types of data processing. In France, for example, the automated analysis of legal decisions including personally identifiable data of judges or court clerks is now a crime, potentially carrying a prison sentence \cite{%
Tashea_2019_France_bans_publishing_of_judicial_analytics_and_prompts_criminal_penalty}.

\subsection{Privacy}
\label{sec:privacy}

While privacy may seem like a fundamental right, %
\e{it is not necessarily clearly defined in national legislation}, complicating its interpretation and implementation by researchers. 
In the US, e.g., current legal provisions are based on a mix of constitutional amendments, federal and state laws, and Supreme Court decisions that are ``patchy and in critical ways outdated'' \cite{CateCate_2012_Supreme_Court_and_information_privacy}. For example, Facebook recently lost its appeal \cite{2021_Facebook_Inc_Petitioner_v_Perrin_Aikens_Davis_et_al} in a case concerning surreptitious tracking of users logged out of Facebook, on the basis of two separate laws: the federal Wiretap Act and California state privacy laws. Most recently in Australia, Facebook lost another case where it was found that they can be held liable for defamation for anything that they post.\footnote{\url{eresources.hcourt.gov.au/downloadPdf/2021/HCA/27}} The landscape is a moving feast for lawyers, let alone for computer scientists and practitioners trying to comply with best practice.

The best-known legislation, setting the current standard for best practice globally regarding data, is the EU General Data Protection Regulation\footnote{\url{eur-lex.europa.eu/legal-content/EN/TXT/HTML/?uri=CELEX:32016R0679}} (GDPR). %
GDPR governs data processing in all EU member states (and the UK, since it has adopted a copy of GDPR \cite{TheNationalArchives_2019_Data_Protection_Privacy_and_Electronic_Communications_Amendments_etc_EU_Exit_Regulations_2019}), as well as processing of personal data of any subjects of EU/UK, \textit{irrespective of where in the world it takes place}. This makes GDPR applicable to most large-scale NLP resources based on web-crawled and social media data, since they are likely to contain at least some samples of data of EU/UK subjects.\footnote{\e{The degree to which extraterritorial laws (including GDPR) can be enforced is debated \cite{Greze_2019_extra-territorial_enforcement_of_GDPR_genuine_issue_and_quest_for_alternatives}, but: (a) their violation does already pose a threat of reputational damage \cite{Azzi_2018_Challenges_Faced_by_Extraterritorial_Scope_of_General_Data_Protection_Regulation}; and (b) legislation in this sphere is quickly developing, and so it is possible that new mechanisms to enforce such laws e.g. through international cooperation may emerge. This might be the reason why, as of January 2020, an estimated 25\% of Fortune 500 US-based retailers simply geo-blocked EU users \cite{2020_What_percentage_of_United_States_retailers_have_decided_to_block_European_visitors_from_reaching_their_websites}}.}

\begin{figure}[t]
    \centering
\begin{mdframed}
\small
Personal data shall be:
\begin{enumerate}[(a)]
    \item processed lawfully, fairly and in a transparent manner in relation to the data subject (\textbf{`lawfulness, fairness and transparency'});
    \item collected for specified, explicit and legitimate purposes and not further processed in a manner that is incompatible with those purposes; further processing for archiving purposes in the public interest, scientific or historical research purposes or statistical purposes shall, in accordance with Article 89(1), not be considered to be incompatible with the initial purposes (\textbf{`purpose limitation'});
    \item adequate, relevant and limited to what is necessary in relation to the purposes for which they are processed (\textbf{`data minimisation'});
    \item accurate and, where necessary, kept up to date; every reasonable step must be taken to ensure that personal data that are inaccurate, having regard to the purposes for which they are processed, are erased or rectified without delay (\textbf{`accuracy'});
    \item kept in a form which permits identification of data subjects for no longer than is necessary for the purposes for which the personal data are processed [...] %
    (\textbf{`storage limitation'});
    \item processed in a manner that ensures appropriate security of the personal data, including protection against unauthorised or unlawful processing [...] %
    (\textbf{`integrity and confidentiality'}).
\end{enumerate}

\end{mdframed}
    \caption{GDPR principles relating to processing of personal data (Article 5)}
    \label{fig:gdpr-article5}
\end{figure}

It is worth noting that the GDPR regime rests on the right to data protection spelt out in Article 8 of the EU Charter of Fundamental Rights,\footnote{The Charter of Fundamental Rights of the European Union OJ C364/01 (\citeyear{2000_Charter_of_Fundamental_Rights_of_European_Union_2000_OJ_C36401}) became legally binding as part of the constitutional law of the EU when the Treaty of Lisbon entered into effect on 1 December 2009.} and so extends well beyond data privacy to human rights. We cannot do this even bigger issue justice within the scope of this paper, but some of the underlying principles are represented in the issues we discuss, and should frame thinking in governance of NLP.

\Cref{fig:gdpr-article5} shows the main GDPR principles that protect privacy. In a nutshell, people need to know and consent to how and why their personal data is being used. Researchers need to obtain consent for specific uses of data, not use it for anything else, not keep it for longer than necessary for the stated purpose, and prevent access to the data by anyone who has not obtained the same consent.

These limitations apply to anything that counts as ``personal data'', defined as ``any information relating to an identified or identifiable natural person (`data subject')''. People can be identified both directly (via name, ID numbers, online identifiers, IP addresses, or location data), or indirectly through ``one or more factors specific to the physical, physiological, genetic, mental, economic, cultural or social identity of that natural person''. 

The GDPR defines pseudo-anonymisation as ``processing of personal data in such a manner that the personal data can no longer be attributed to a specific data subject without the use of additional information, provided that such additional information is kept separately'' so as to prevent (re\babelhyphen{nobreak})identification. 
Anonymisation techniques are an active research area, but at present ``even coarse datasets provide little anonymity'' %
\cite{deMontjoyeHidalgoEtAl_2013_Unique_in_Crowd_privacy_bounds_of_human_mobility}. There is also a fundamental tradeoff between data utility and the privacy it affords: any anonymisation technique necessarily removes information from data, which may significantly decrease its utility \cite{BrickellShmatikov_2008_cost_of_privacy_destruction_of_data-mining_utility_in_anonymized_data_publishing,ZangBolot_2011_Anonymization_of_location_data_does_not_work_large-scale_measurement_study}. 

Some argue that de-identified data needs to be better defined and regulated, as much data that is currently defined as de-identified is easily reidentifiable \cite{CulnaneLeins_2019_Misconceptions_in_privacy_protection_and_regulation}. In NLP in particular, we have the fundamental problem that longer excerpts of text are almost impossible to anonymise (even in the limited sense of ``masking entity references'' \cite{MozesKleinberg_2021_No_Intruder_no_Validity_Evaluation_Criteria_for_Privacy-Preserving_Text_Anonymization}). De-identification may happen in two ways: %

\begin{compactitem}
    \item \textit{Direct identification:} the authorship of the text or message may be stripped from the resource, but publicly available elsewhere (through the search function on a social media platform, or generic web search). %
    \e{Combined with metadata (such as that the utterance in question was used on Yelp), even relatively small amounts of texts may potentially be uniquely identifying \cite{shrestha-etal-2017-convolutional,kestemont2019overview}.}
    \item \textit{Indirect identification} through authorship attribution: it may also be possible to identify the author of non-public texts based on other texts of theirs that \textit{are} public. For example, anonymous social media users can be identified based on stylometry and typo patterns \cite{NarayananPaskovEtAl_2012_On_Feasibility_of_Internet-Scale_Author_Identification}.
\end{compactitem}

\e{In some NLP applications such as search or voice assistants, we can mitigate potential re-anonymisation through differential privacy \cite{DworkMcSherryEtAl_2006_Calibrating_Noise_to_Sensitivity_in_Private_Data_Analysis}, where the individual data points are deliberately superimposed with noise or include only relatively frequent data points using cardinality estimation \cite{HarmouchNaumann_2017_Cardinality_estimation_experimental_survey}. But large language models can and do memorise the training texts \cite{CarliniTramerEtAl_2020_Extracting_Training_Data_from_Large_Language_Models}, and training them with privacy-preserving techniques without sacrificing too much quality is an active research area \cite{BasuRoyEtAl_2021_Benchmarking_Differential_Privacy_and_Federated_Learning_for_BERT_Models}. If a language model reproduces a memorised excerpt, and the original author is easy to identify via web search, the author could be exposed to unwelcome attention (similarly to how an out-of-context quote can inflict serious reputational damage).}

\subsection{Transparency}
\label{sec:transparency}

Transparency ``in relation to data subject'' is the first GDPR principle cited in \Cref{fig:gdpr-article5}: the people whose data is being used need to know exactly what was collected, and what it will be used for. It is transparency that prevents the collected data from simply being reused for other purposes to which the data subjects would not have consented.

Article 12 stipulates that the data subjects need to be informed through ``a concise, transparent, intelligible and easily accessible form, using clear and plain language''. In particular, according to Articles 12--14, the data subject must be aware of the identity and contact details of the controller, the purposes of the processing and its legal basis, the recipients of the data, and the period of storage. They must also be made aware of their rights, including the right to request rectification or erasure of the data, request its copy, impose restrictions on its processing, or even withdraw consent. 

Importantly for NLP, if the application involves any ``automated decision-making, including profiling'', the data subjects have to be provided with ``meaningful information about the logic involved, as well as the significance and the envisaged consequences of such processing for the data subject'' (Article 12).

\subsection{Reproducibility}

The reproducibility principle stems from general scientific methodology. It requires that researchers focus on the more robust observations, and guards against falsification and fabrication (as such results would not be reproducible). Most experiments and studies are performed once, under unique conditions, which means that it is difficult to guarantee that the results are valid and trustworthy. \e{In NLP, to reproduce an experiment one would need both the implementation of a given system, and the data on which it would run.}%

The increased focus on \textit{code availability} in NLP research \cite[][inter alia]{WielingRaweeEtAl_2018_Squib_Reproducibility_in_Computational_Linguistics_Are_We_Willing_to_Share,Raff_2019_Step_Toward_Quantifying_Independently_Reproducible_Machine_Learning_Research,DodgeGururanganEtAl_2019_Show_Your_Work_Improved_Reporting_of_Experimental_Results,Crane_2018_Questionable_Answers_in_Question_Answering_Research_Reproducibility_and_Variability_of_Published_Results} leads to increased expectations for \textit{data availability}.\footnote{In addition to lower-level issues such as fixed splits of the data, the choice of evaluation metric(s), and potentially even pre-processing strategies.} Two recent taxonomies of reproducibility in NLP \cite{CohenXiaEtAl_2018_Three_Dimensions_of_Reproducibility_in_Natural_Language_Processing,TatmanVanderPlasEtAl_2018_Practical_Taxonomy_of_Reproducibility_for_Machine_Learning_Research} both consider data availability a necessary precondition. \e{This is completely fair when the data is e.g. synthetic or collected with informed consent for this use, but in many other cases the principle of reproducibility is inherently in tension with the other principles, as will be discussed in \Cref{sec:tension}.} %

\subsection{Do No Harm}

The final principle comes from work on ethical AI and the social impact of NLP. While the NLP community ideally aims to improve the world, the real world is far more complex than the binary categories of `good' and `bad' \cite{Green_2019_Good_isnt_good_enough}. %
Context and culture play a role, with no one-size-fits-all solution. Inclusive, representative groups on projects and reviews are part of the solution, but not enough on their own.

\e{When we think of ``do no harm'', the first association may be harm inflicted by deployed systems or direct experiments on users \cite{KramerGuilloryEtAl_2014_Experimental_evidence_of_massive-scale_emotional_contagion_through_social_networks,Flick_2016_Informed_consent_and_Facebook_emotional_manipulation_study}, but there are also best-practice recommendations for data collection:}

\begin{compactitem}
    \item \textit{Document the population from whom the data comes.} \e{This is necessary both for understanding the linguistic data, if the study goals are descriptive, and for ensuring a match with the target user demographics, if the data or models derived from it are meant for real-world use \cite{Rogers_2021_Changing_World_by_Changing_Data}}. Much of NLP data is ``convenience samples'' of naturally-occurring data, which reflects a world riddled with inequalities and social biases. For instance, GPT-2 training data was scraped from links shared on Reddit, reflecting the worldview, language, and interests of predominantly young white men \cite{BenderGebruEtAl_2021_On_Dangers_of_Stochastic_Parrots_Can_Language_Models_Be_Too_Big}. Creation of `perfect' and representative samples is not necessarily possible, but documenting the lacunae and omissions is \cite{BenderFriedman_2018_Data_Statements_for_Natural_Language_Processing_Toward_Mitigating_System_Bias_and_Enabling_Better_Science}.
    \item \textit{Consider the potential for exposure.} Social media storms are a force to be reckoned with, which can equally serve as an accountability instrument for public figures and companies \cite{RostStahelEtAl_2016_Digital_Social_Norm_Enforcement_Online_Firestorms_in_Social_Media,NeuSaxtonEtAl_2019_Twitter_and_social_accountability_Reactions_to_Panama_Papers}, or a means of identity-based harrassment \cite{Ortiz_2020_Trolling_as_Collective_Form_of_Harassment_Inductive_Study_of_How_Online_Users_Understand_Trolling,Waisbord_2020_Mob_Censorship_Online_Harassment_of_US_Journalists_in_Times_of_Digital_Hate_and_Populism}. 
    Taking data from a moment in time and then `baking it in' presents many problems, potentially leading to inaccurate or harmful representations of the individual (e.g.\ if they later retract a comment, or appeal a legal case and have the decision reversed).
    \item \textit{Consider the potential for misuse.} The responsible thing to do before commencing a project is to think through what a bad actor could do with its results. *ACL conferences are increasingly requiring all submissions %
    to consider possible misuse\footnote{\url{2021.naacl.org/ethics/faq}} and how the possible harms should be mitigated. In some cases it may not be safe to release even the annotation guidelines, much less the data \cite{RogersKovalevaEtAl_2019_Calls_to_Action_on_Social_Media_Potential_for_Censorship_and_Social_Impact}.
\end{compactitem}

\section{Tension between Data Collection Principles}
\label{sec:tension}

While all of the above principles are important, there is a tug-of-war between them, \e{as well as tensions with other factors in the research environment}. The reproducibility principle aims to maximise data sharing in the interest of open science, while the others all limit it from ethical and legal perspectives. The transparency principle means that the data processors have to disclose what they are doing with the data, but that complicates the protection of trade secrets. The privacy principle is sometimes fundamentally at odds with public interest (crime, harassment, and accountability for public figures and organisations). Reproducibility and preservation of records of public interest may be in conflict with data subject privacy, the right to be forgotten, client confidentiality, and legitimate business objectives of industry research. 

ToS of individual platforms may further cross-cut these principles: e.g.\ Twitter's requirement that only tweet IDs may be distributed maximises user control over their data, but it makes anonymisation impossible, and also sacrifices reproducibility (due to data attrition). Add to that the differences in national legislation, cultural norms, and priorities of individual researchers, and it is evident that there are no clear solutions that simply follow from all the above principles.

The result is that different research communities have come up with different norms and combinations of these principles, depending on their agenda, the power differential of data controllers and data subjects, and simply the goodwill of researchers willing to volunteer time to engage in unheralded administrative and advocacy work to change the norms. 

As an example, consider the tension between privacy and open science in medical research \cite{MinssenRajamEtAl_2020_Clinical_trial_data_transparency_and_GDPR_compliance_Implications_for_data_sharing_and_open_innovation}. Fundamentally, patient data has to be kept private, and most clinical studies do not publish it. This, however, prevents joint analysis of data from independent clinical trials, which could yield critical insights and literally save lives of other patients. One solution is adaptive clinical trial platforms \cite{AngusAlexanderEtAl_2019_Adaptive_platform_trials_definition_design_conduct_and_reporting_considerations}, which enable cross-institutional coordination and data sharing in ongoing trials. But currently such initiatives rely largely on volunteer service work by researchers \cite{HickeyChen_How_to_Fix_Incentives_in_Cancer_Research_Ep_449}. The result is patchy: for example, there is a platform for consolidating pancreatic cancer research across institutions,\footnote{\url{www.pancan.org}} but not for lung cancer.\footnote{\url{www.lung.org/research/clinical-trials}} %

NLP is not an exception: as an interdisciplinary field, it has a mixture of players with different priorities and incentive structures, who argue for different combinations of the above principles, in part predicted by the domain they work in (e.g.\ clinical NLP with a strong focus on privacy and transparency vs.\ machine learning for NLP with a strong focus on reproducibility). More recently, as part of the rush to develop and distribute pre-trained language models, there has arguably been an over-focus on data volume and too little focus on any of the data principles, including the use of corpora of dubious legality and quality such as BookCorpus \cite{BandyVincent_2021_Addressing_Documentation_Debt_in_Machine_Learning_Research_Retrospective_Datasheet_for_BookCorpus}.  %

\begin{figure*}[t!]
\begin{mdframed}
\centering
\small

\textbf{For papers using a previously-published resource:}

\begin{checklist} %
    \item \e{The authors explain their choice of data, given the available resources and their known limitations (e.g. representativeness issues, biases, annotation artifacts) and any data protection issues (e.g. inclusion of sensitive health data).} \e{\textit{\hfill See Section \_\_\_}}.
    \item The authors discuss whether their use of a previously-published resource is compatible with its original purpose and license, and any known limitations (e.g. if the target user group is represented in the sample). \textit{\hfill See Section \_\_\_}.
\end{checklist}

\textbf{For papers contributing a new resource:}

\begin{checklist}

    \item The authors have the \textbf{legal basis} for processing the data and, if it is made public, for distributing it. \textit{(Check one)}
    \begin{checklist}
    \item The data is in public domain, and licensed for research purposes;
    \item The data is used with consent %
    of its creators or copyright holders;
    \item If the data is used without consent, the paper makes the case to justify its legal basis %
    (e.g. research performed in the public interest under GDPR). \textit{\hfill See Section \_\_\_}.
    \end{checklist}

    \item The paper describes in detail the \textbf{full data collection protocol}, including collection, annotation, pre-processing, and filtering procedures. In the case that the dataset involves work by human subjects (e.g.\ data creation or annotation), the paper describes efforts to ensure fair compensation. \textit{\hfill See Section \_\_\_}.
    
    \item \textbf{Safe use} of data is ensured. \textit{(Check all that apply)}
    \begin{checklist}
    \item The data does not include any protected information %
    (e.g. sexual orientation or political views under GDPR), or a specified exception applies. \textit{\hfill See Section \_\_\_}.
    \item The paper \e{is accompanied by} a data statement describing the basic demographic and geographic characteristics of the population that is the source of the language data, and the population that it is intended to represent. \\ \textit{\hfill See \_\_\_}.
    \item If applicable: the paper describes whether any characteristics of the human subjects were self-reported (preferably) or inferred (in what way), justifying the methodology and choice of description categories. \textit{\hfill See Section \_\_\_}.
    \item The paper discusses the harms that may ensue from the limitations of the data collection methodology, especially concerning marginalized/vulnerable populations, and specifies the scope within which the data can be used safely. \textit{\hfill See Section \_\_\_}.
    \item If any personal data is used: the paper specifies the standards applied for its storage and processing, and any anonymization efforts. \textit{\hfill See Section \_\_\_}.
    \item If the individual speakers remain identifiable via search: the paper discusses possible harms from misuse of this data, and their mitigation. \textit{\hfill See Section \_\_\_}.
    \end{checklist}
    
    \item If any data or models are made public: \textbf{safe reuse} is ensured. \textit{(Check all that apply)}
    \begin{checklist}
    \item The data and/or pretrained models are released under a \e{specified} license that is compatible with the conditions under which access to data was granted (in particular, derivatives of data accessed for research purposes should \textit{not} be deployed in the real world as anything other than a research prototype, especially commercially). \textit{\hfill See \_\_\_}.
    \item The paper specifies the efforts to limit the potential use to circumstances in which the data/models could be used safely (such as an accompanying data/model statements). \textit{\hfill See Section \_\_\_}.
    \end{checklist}

\item The data collection protocol was \textbf{approved by the ethics review board} at the authors' institution, or such review is not applicable for specified reasons. \textit{\hfill See Section \_\_\_}.    
\end{checklist}

\end{mdframed}
    \caption{Responsible Data Use Checklist}
    \label{fig:checklist}
\end{figure*}

\section{Responsible Data Use Checklist}
The conflicting principles of data collection, coupled with conflicting priorities of different researchers, make for a torn, confused community. The movement towards standardising ethical and legal expectations has already started: %
NeurIPS 2020 pioneered obligatory broader impact statements inspired by \citet{2018_Its_Time_to_Do_Something_Mitigating_Negative_Impacts_of_Computing_Through_Change_to_Peer_Review_Process}, %
and NAACL, ACL-IJCNLP and EMNLP 2021 adopted a similar ethics policy, with a separate committee ethically reviewing papers flagged by regular reviewers. In parallel with all that, the *ACL conferences now use an adapted version of the reproducibility checklist \cite{DodgeSmith_2020_Guest_Post_Reproducibility_at_EMNLP_2020}. %

In this paper, we are hoping to refine the conversation regarding responsible data use, \e{collection and distribution}, as one thread of a richer dialog about NLP ethics.
To this end, we propose to re-structure the host of issues that the paper authors are asked to consider into three areas: 

\begin{compactitem}
    \item \textit{experiment reproducibility}: provision of code, distribution of trained models/model outputs, specification of hyper-parameters, standardisation of resource usage, etc.
    \item \textit{broader impacts of the NLP task/system}: could the research be (mis)used\footnote{\e{A frequent counter-argument to consideration of broader impacts of NLP research is that anything could be misused, even ``neutral'' tools like parsers. While that is true, the ease, likelihood, and probable consequences of misuse do matter. NLP is in dire need of research into the cost--benefit of the use of its systems \cite{Rogers_2021_Changing_World_by_Changing_Data}.}} to amplify existing social inequalities, support/undermine democratic processes, \e{be} weaponised, \e{be} retooled for propaganda purposes, increase the likelihood of armed conflicts? Are there potential conflicts of interest given the funding sources? \e{Are there significant carbon costs?} \e{For more discussion of these (and many other) issues see} \citet{AI_2020_Guide_to_Writing_NeurIPS_Impact_Statement}, \citet{2018_Its_Time_to_Do_Something_Mitigating_Negative_Impacts_of_Computing_Through_Change_to_Peer_Review_Process}, and the NAACL 2021 Ethics FAQ %
    \item \textit{responsible data use/reuse}: whether the collection methodology is sufficiently described, complies with applicable regulations, and conditions for safe use/reuse are specified.
\end{compactitem}

\e{This paper focuses on the third area. Inspired by  \citet{DodgeSmith_2020_Guest_Post_Reproducibility_at_EMNLP_2020}, we make a %
first attempt at a checklist for assessing responsible data (re-)use, shown in \Cref{fig:checklist}. Similarly to the reproducibility checklist, it is voluntarily filled in by the authors of an NLP study. Ideally it would accompany the papers as an appendix section \textit{both pre- and post-publication}. This way it would not only facilitate peer review, but also help the regular readers of the paper to determine if \textit{they} can use the proposed model or data, given the rules of their institution. If multiple datasets are presented or used, each one would have a separate checklist.}

Our proposal draws on the GDPR framework, NAACL ethics guidelines, and work on documenting the populations represented in NLP resources \cite{BenderFriedman_2018_Data_Statements_for_Natural_Language_Processing_Toward_Mitigating_System_Bias_and_Enabling_Better_Science}. While by no means a complete solution, we believe that it is a useful starting point. %

The main section of the checklist focuses on the \textbf{safe use} of data, with many questions overlapping with those expected in institutional ethics review (see \Cref{sec:irb}). It also has a section dedicated to \textbf{safe reuse}, which aims to \e{nudge} the resource authors towards specifying limitations and safe use cases. At the same time, the model developers are nudged towards specifying whether their use is consistent with any such limitations.

\e{A key provision of safe resuse is} \textbf{ensuring consistent terms of data sharing}. %
In many cases researchers are allowed access to data that would not be granted to commercial entities, under the provision that this is done solely for research purposes, or for `social good'. \e{For example, based on Article 89 of GDPR and their local legislation, EU-based researchers may be exempted from observing some data subject rights ``for scientific or historical research purposes or statistical purposes'' and research performed ``in public interest". If researchers gain access to data under such provisions, and then publish the data (or models derived from that data) under licenses allowing commercial use, this creates a loophole.}

\e{While we argue that the community should aim for a common standard, our checklist of course does not fully specify it: what data is considered sensitive, what payment fair, what legal grounds acceptable for processing data without data subject consent? What it does achieve is forcing the authors to explicitly consider all these issues and present their motivation for the choices they made. If they went through a thorough review by a local board, they would already have done this work, and so workload should not increase.\footnote{\e{The added bonus is that once the authors of a resource document their decisions this way, it will be easier for others to motivate their use of that resource, providing an indirect incentive for the dataset creators.}}

We readily acknowledge that the development of a shared norm for what is and is not responsible data use is already happening. That process is the reason why from time to time the choices made in a given paper provoke heated discussion, which has impact on the planning of further projects by the community. We admit that even with the best checklists the result will likely never be perfect, as it is a reflection of our ever-changing field. The difference is that a standardised practice of showing the legal and ethical reasoning behind different projects in a structured way should accelerate the evolution of the shared norm, and decrease misunderstandings. %
Also, in the case of difficult legal disputes such as disagreements in ToS interpretation, a professional organisation such as ACL could consult legal professionals, and build a repository of common issues for the community to learn from.}

\section{Can't We Just Defer to IRB?}
\label{sec:irb}

\e{Our proposed checklist has a separate checkbox for whether the project was reviewed by an ethics review board at the researchers' home institution. In the US it would be an Institutional Review Board (IRB).\footnote{The names and processes of such boards differ across countries and institutions: Comit\'e de Protection des Personnes (France), Human Research Ethics Committee (Australia), Institutional Ethics Committee (India), Research Ethics Committee (UK).} Ideally, such a board would pre-check everything discussed in this paper, \textit{before} the project is carried out. We absolutely encourage NLP researchers to use the legal and ethical guidance their institutions provide. 

However, the primary goal of ethics review boards is \textit{planning} of research projects. %
We argue that the ethical and legal thinking behind a project should be systematically made available for the community \textit{post-publication}, so that future researchers can make better decisions about what they reuse, more easily document the limitations they inherit, and learn from precedents when developing the ethical and legal motivation for new projects. Reading papers in search of resources and models we can use, given our local rules and legislation, would be much easier if there were a standard way to share such information. In our proposal it is a checklist included in the appendix, with blank fields for specifying the sections of the paper where the given information can be found.

In the context of peer review, we also argue that a professional organisation such as ACL has a role in establishing a set of consistent \textit{minimal} expectations across the NLP community (while the authors are also expected to comply with their national legislation and institutional rules). This is necessary for the following reasons: %

\begin{compactenum}[(1)]
    \item \e{There is a lot of variation in the researcher perception of when ethics review is needed \cite{ShmueliFellEtAl_2021_Beyond_Fair_Pay_Ethical_Implications_of_NLP_Crowdsourcing}. \citet{SantyRaniEtAl_2021_Use_of_Formal_Ethical_Reviews_in_NLP_Literature_Historical_Trends_and_Current_Practices} showed that less than 0.8\% of NLP studies published since 2006 have sought IRB approval, and that was mostly for data collection or annotation. Approvals are rarely sought for data scraping or re-purposing, or for systems. Clearly, more than 1\% of NLP papers come from US institutions and have something to do with data collection.
    \item There are major differences in both national legislation and institutional practices. %
    In the worst-case scenario, without a minimal set of expectations, an institution could perform research disregarding privacy or human rights, and ACL would have no basis to block its publication.} %
\item Some current review boards do not have the legal and ethical expertise to review legal compliance, and their evaluation may thus be less rigorous. \e{The target scope of their scrutiny also varies, but the average reader or reviewer of the paper may not necessarily know what the review by committee type X entails. E.g. in the debate after NAACL 2021 business meeting several researchers expressed the belief that the ACL ethics review is superfulous because US IRBs cover all the target issues. In fact they are \textit{discouraged} from considering long-term impact of research \cite{FleischmanLevineEtAl_2011_Dealing_With_Long-Term_Social_Implications_of_Research}.} 
\item The ethics review process is bound to produce cases where the authors and reviewers rely on different sets of rules, some more stringent than others\footnote{\url{twitter.com/zehavoc/status/1430793222150840322}}. The authors should generally be able to argue for their ``home'' rules, but only as long as they do not go below the minimal standard shared by the community -- else we are in the situation (2) above. 
\end{compactenum}
}

\section{Challenges}

The above checklist is by no means exhaustive in terms of legal and ethical considerations in NLP projects, and does not solve all the problems. In particular, the following topics require a lot more legal thinking in the community. 

\paragraph{Web crawled data} NLP models are often pre-trained on large volumes of web crawled text: from 6B words of news data in word2vec \cite{MikolovChenEtAl_2013_Efficient_estimation_of_word_representations_in_vector_space} to 42B words of Common Crawl data in GloVe \cite{PenningtonSocherEtAl_2014_Glove_Global_vectors_for_word_representation}, and now 300B tokens in GPT3 \cite{BrownMannEtAl_2020_Language_Models_are_Few-Shot_Learners}, and counting \cite{FedusZophEtAl_2021_Switch_Transformers_Scaling_to_Trillion_Parameter_Models_with_Simple_and_Efficient_Sparsity}. These studies, like any other, have to have a legal basis for data collection, storage, and processing.

GDPR recognises ``public interest'' as a legal basis for processing of personal data (in this case, data attributable to individuals through public records of authorship). But most current large-scale models are trained by industry labs and made available for commercial use. %
The only possible basis for such training is the business' ``legitimate interest'', but then it needs to outweigh the interests of the data subjects, who may suffer significant harms \cite{BenderGebruEtAl_2021_On_Dangers_of_Stochastic_Parrots_Can_Language_Models_Be_Too_Big} %
Furthermore, those involved in data storage and processing still have to comply with the privacy principles, including the protection of personal data. There is ongoing work in this direction in the Big Science Workshop.\footnote{\url{bigscience.huggingface.co}}

All the above only concerns privacy legislation, which copyright law is orthogonal to. %
Copyright would be directly violated by the act of copying the data for training, and also potentially by the fact of storing the protected text in trained models. With static word embeddings, it could be argued that the model itself does not ``store'' any text, but the current generation of language models clearly do reconstruct excerpts of training data. Since this is a blackbox process, it is hard to tell how much ``creativity'' there is in any particular sample, but the worst-case scenario is that such models could make their users potentially liable for plagiarism.

\paragraph{Legacy data} Many datasets that are currently in use have been released with little consideration for copyright, speaker demographics, or proper handling of personal information \cite{BandyVincent_2021_Addressing_Documentation_Debt_in_Machine_Learning_Research_Retrospective_Datasheet_for_BookCorpus}. Since such resources have been published in prestigious venues, some NLP researchers might feel that there is precedent to use them and build new resources in a similar way. %

The ongoing use of such data raises many questions. It is not realistic to prohibit its use overnight, but, if a responsible data-use standard is established, it should give rise to a new generation of resources that do not have such problems. In the short term, such a standard should also stimulate efforts to address documentation debt \cite{BandyVincent_2021_Addressing_Documentation_Debt_in_Machine_Learning_Research_Retrospective_Datasheet_for_BookCorpus,DodgeSapEtAl_2021_Documenting_English_Colossal_Clean_Crawled_Corpus}, as well as analysis and cleaning of the older resources (similarly to the problem of removal of pornographic images that the computer vision community is tackling\footnote{See \url{www.losinglena.com}, for example.}). Note that our checklist includes a section for previously-published resources, which should nudge authors towards more carefully documented data (since it would be easier to motivate).

\paragraph{Consent}
The key provision of use for data protected by both privacy and copyright legislation is that any processing can only happen with consent of the data subject/copyright owner. \e{\citet{FieslerProferes_2018_Participant_Perceptions_of_Twitter_Research_Ethics} report that regular Twitter users are likely to not even be aware that their tweets can be used by researchers, and the willingness for their data to be analysed depends on the kind of study. %
GDPR offers some guidance on} how data subjects need to be informed of what data was collected and what for, but there are other questions for which we do not have clear answers. In particular, should the participants be made aware of the many ways in which they can be identified in anonymised data (see \Cref{sec:privacy})? What constitutes ``public'' data, and does the fact that someone willingly made some text public on their website or social media page constitute tacit permission to use it for any kind of research (or training commercial models)? What about the frequent situation of data that is originally released legitimately, but subsequently withdrawn? %
Is it permissible to use leaked data? If research is done without consent in the public interest, how can we assess the claims of its benefits vs.\ possible harms?

We hope that this paper stimulates further discussion of these and other issues.

\section{Conclusion}

As NLP technologies become more mainstream and commercialised, it is increasingly important that the NLP community leads best practice in legal and ethical standards.  This paper is a step in that direction. We propose a tentative checklist %
for responsible data (re-)use, intended to serve as a complement to the existing reproducibility checklist. %
We hope that the community will continue working towards a common standard for data collection and sharing, complemented with transparent review mechanisms to ensure that the standard is upheld. %

\section{Acknowledgements}

This paper owes a lot to Emily M.~Bender, Catalina Goanta, Robert (Munro) Monarch, Anna Rumshisky, and numerous discussions in the \#NLProc Twitter community. We also thank the anonymous reviewers for their insightful comments.

\bibliographystyle{acl_natbib}
\bibliography{anthology,acl2021}

\end{document}